\documentclass[10pt,journal,compsoc]{IEEEtran}

\ifCLASSOPTIONcompsoc
\usepackage[nocompress]{cite}
\else
\usepackage{cite}
\fi
\usepackage{graphicx} 
\usepackage{subfigure}

\usepackage{epsfig}
\usepackage{amsmath}
\usepackage{amssymb}
\usepackage{algorithm}
\usepackage{algorithmic}
\usepackage{url}

\usepackage[export]{adjustbox}

\usepackage{color}

\usepackage[pagebackref=true,breaklinks=true,letterpaper=true,colorlinks,bookmarks=false]{hyperref}
\newcommand{\ignorethis } [1] {}

\newcommand{\etal       }     {{et~al.}}

\newcommand{\eg         }     {{e.g.}}

\newcommand{\Reals      }     {{\textrm{I\kern-0.18em R}}}

\newcommand{\change     } [1] {\mbox{{\footnotesize $\Delta$} \kern-3pt}#1}

\newcommand{\Norm       } [1] {{\left\| #1 \right\|}}

\newcommand{\Pnorm      } [2] {\Norm{#1}_{#2}}

\definecolor{darkred}{rgb}{0.7,0.1,0.1}
\definecolor{darkgreen}{rgb}{0.1,0.7,0.1}
\definecolor{cyan}{rgb}{0.7,0.0,0.7}
\definecolor{dblue}{rgb}{0.2,0.2,0.8}
\definecolor{maroon}{rgb}{0.76,.13,.28}
\definecolor{burntorange}{rgb}{0.81,.33,0}
\definecolor{white}{rgb}{1,1,1}

\def\ShowNotes{}

\ifdefined\ShowNotes
  \newcommand{\colornote}[3]{{\color{#1}\bf{#2 #3}\normalfont}}
\else
  \newcommand{\colornote}[3]{}
\fi

\newcommand {\dcor}[1]{\colornote{red}{DC:}{#1}}
\newcommand {\dov}[1]{\colornote{blue}{DD:}{#1}}

\newcommand {\whitetxt}[1]{\colornote{white}{}{#1}}

\newcommand\todosilent[1]{}

\ifdefined\SmallImages
  
\else
  
\fi

\newcommand{\shortcite}[1]{\cite{#1}}

\newcommand{\projurl}{http://faces.cs.princeton.edu/}

\makeatletter
\newcommand\footnoteref[1]{\protected@xdef\@thefnmark{\ref{#1}}\@footnotemark}
\makeatother

\usepackage{cleveref}

\definecolor{blue}{rgb}{0,0,1}
\definecolor{red}{rgb}{1,0,0}
\definecolor{orange}{rgb}{0.75, 0.4, 0}

\usepackage{booktabs}

\usepackage[font=small]{caption}
\begin{document}
	
	\title{Unsupervised Natural Image Patch Learning}
	
	\author{Dov~Danon,~\IEEEmembership{}
		Hadar~Averbuch-Elor,~\IEEEmembership{}
		Ohad Fried,~\IEEEmembership{}
		Daniel~Cohen-Or~\IEEEmembership{}}
		


	\IEEEtitleabstractindextext{%

\begin{abstract}
Learning a metric of natural image patches is an important tool for analyzing images. An efficient means is to train a deep network to map an image patch to a vector space, in which the Euclidean distance reflects patch similarity. 
Previous attempts learned such an embedding in a supervised manner, requiring the availability of many annotated images.
In this paper, we present an unsupervised embedding of natural image patches, avoiding the need for annotated images. The key idea is that the similarity of two patches can be learned from the prevalence of their spatial proximity in natural images. 
Clearly, relying on this simple principle, many spatially nearby pairs are outliers, however, as we show, the outliers do not harm the convergence of the metric learning. We show that our unsupervised embedding approach is more effective than a supervised one or one that uses deep patch representations. Moreover, we show that it naturally leads itself to an efficient self-supervised domain adaptation technique onto a target domain that contains a common foreground object.
\end{abstract}

		\begin{IEEEkeywords}
			unsupervised learning, metric learning
	\end{IEEEkeywords}}

		\twocolumn[{%
		\renewcommand\twocolumn[1][]{#1}%
		\maketitle
		\begin{center}
			\vspace{-8pt}
			\centering
			\includegraphics[width=\textwidth]{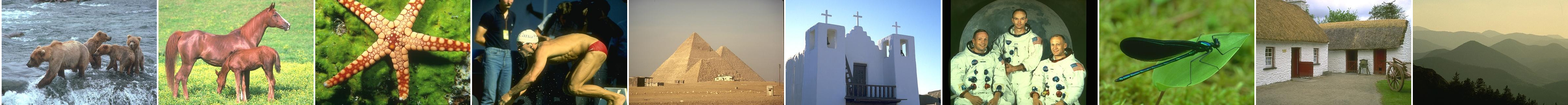}
			\includegraphics[width=\textwidth]{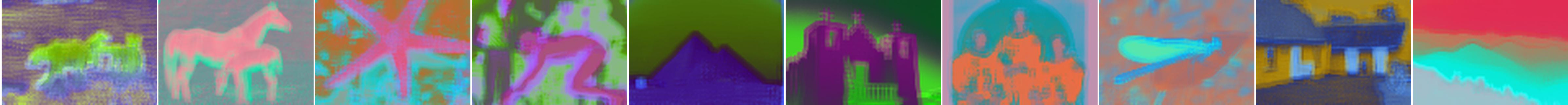} \\
			\vspace{1pt}
			\includegraphics[width=\textwidth]{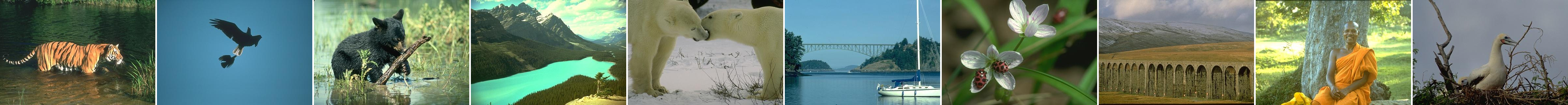}
			\includegraphics[width=\textwidth]{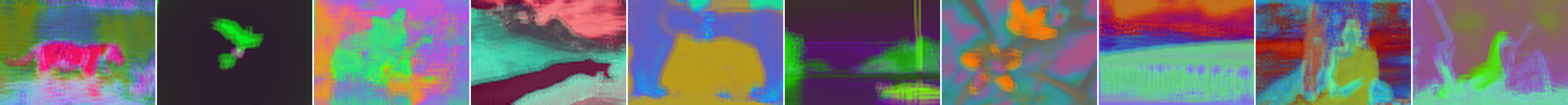}
			\captionof{figure}{Given a natural input image, our technique learns a high-dimensional embedding space, where Euclidean distances between embedded image patches reflect their similarity (visualized in pseudo-RGB colors).}
			\label{fig:teaser}
		\end{center}%
	}]


	\IEEEdisplaynontitleabstractindextext
	
	\IEEEpeerreviewmaketitle

\section{Introduction}

Humans can easily understand what they see at different regions in an image, or tell whether two regions are similar or not.
However, despite 
recent progress,
such forms of image understanding remain extremely challenging. 
One way to address image understanding takes inspiration from the ability of human observers to understand image contents, even when viewing through a small observation window. Image understanding can be formalized as the ability to encode contents of small image patches into representation vectors.
To keep such encodings generic, they are not predetermined by certain classes, but instead aim to project image patches into an embedding space, where Euclidean distances correlate with general similarity among image patches. As natural patches form a low dimensional manifold in the space of patches \cite{matviychuk2015exploring,shi2007mapping}, such an embedding of image patches allows various image understanding and segmentation tasks. For example, semantic segmentation is reduced into a simple clustering technique based on $l_2$ distances. 


The key insight of our work is that such an embedding of image patches can be trained by a neural network in an \emph{unsupervised} manner. Using semantic annotations allows a direct sampling of positive and negative pairs of patches that can be embedded using a triplet loss \cite{Patch2Vec}. However, data labeling is laborious and expensive. Therefore, only a tiny fraction of the images available online can be utilized by supervised techniques, necessarily limiting the learning to a bounded extent.
An unsupervised embedding can also be based on deep patch representations that are learned indirectly by the network, e.g., \cite{doersch2015unsupervised}, however, as we show, explicitly training the network for an embedding can achieve significantly higher performance.

\begin{figure}[t]
	\centering	\includegraphics[width=\columnwidth]{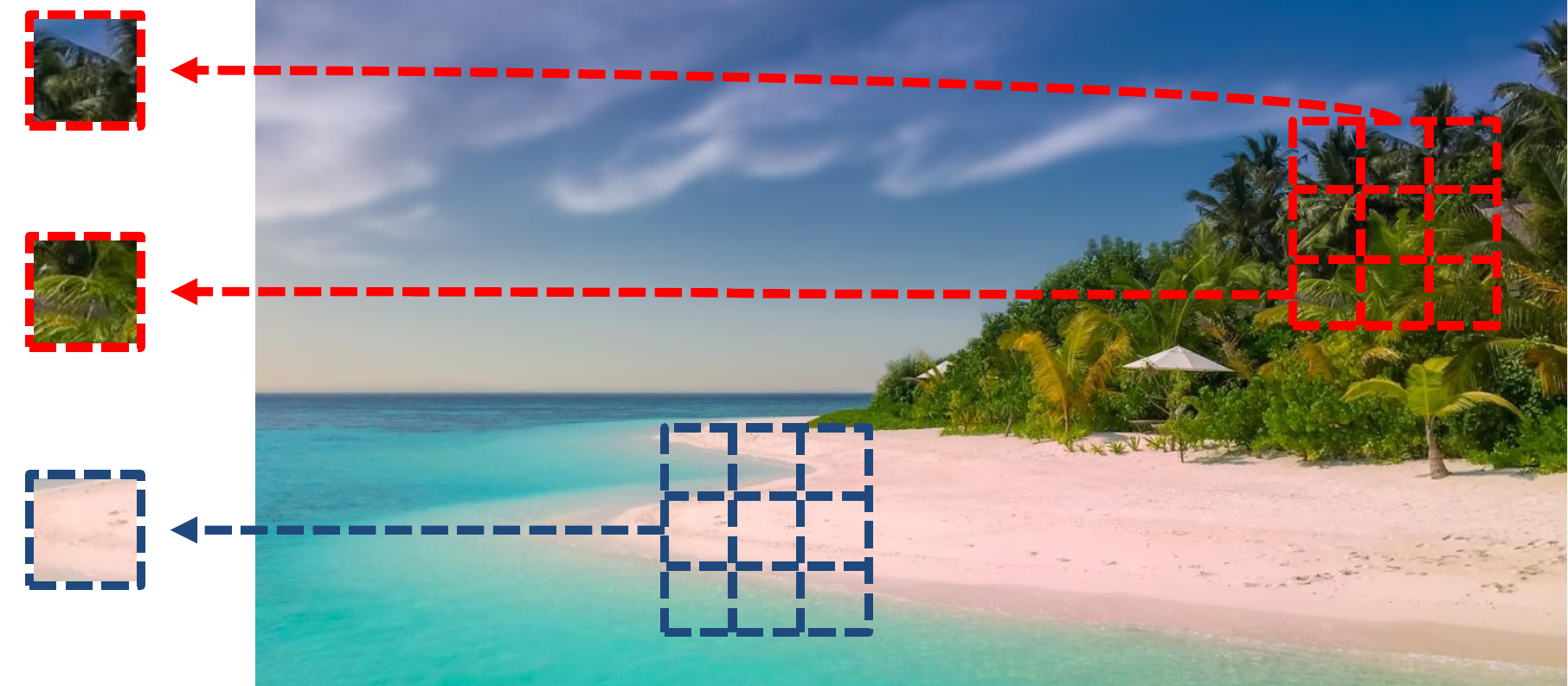}   
	\caption{Learning patch similarity from spatial distances. The premise is that two patches sampled from the same swatch (colored in red) are more likely to be similar to each other than to a patch sampled in a distant one (colored in blue). }
	\label{fig:core_idea}
    \vspace{-8pt}
\end{figure}

In this work, we introduce an unsupervised patch embedding method, which analyses natural image patches to define a mapping from a patch to a vector, such that the Euclidean distance between two vectors reflects their perceptual similarity. We observe that the similarity of two patches in natural images correlates with their spatial distances. In other words, patches of coherent or semantic segments tend to be spatially close, hence forming a  surprisingly simple but strong correlation between patch similarity and spatial distance. Clearly, not all neighboring patches are similar (see~\Cref{fig:core_idea}).
However, as we shall show, these dissimilar close patches are rare enough and uncorrelated, resulting in insignificant noise to the learning system which does not prohibit the learning. 

Our embedding yields \emph{deep images}, as each patch is mapped to a vector of 128D by a deep network. See the visualization of the deep images in the second and fourth rows of ~\Cref{fig:teaser}, obtained by projecting the 128D vectors onto their three principle directions, producing pseudo-RGB images where similar colors correspond to similar embedded points. Using our embedding technique, we further present a domain specialization method. Given a new domain that contains a common foreground object, using self-supervision, we refine the initial embedding results for the specific domain to yield a more accurate embedding.

We use a Convolutional Neural Network (CNN) to learn a 128 dimension embedding space. We train the network on 2.5M natural patches with a triplet-loss objective function. \Cref{sec:method} explains our embedding framework in detail. \Cref{sec:new_domain} describes our domain adaptation technique to a target domain that contains a common foreground object.
In \Cref{sec:results}, we show that the patch embedding space learned using our method is more effective than embedding spaces that were learned with supervision or those based on hand-crafted features or deep patch representations. We further show that by fine-tuning the network to a specific domain using self-supervision, we can further increase performance.


	\section{Related Work}

Our work is closely related to dimensionality reduction and embedding techniques, image patch representation, transfer learning and neural network based optimization. In the following we highlight directly relevant research.


Image patches can be treated as a collection of partial objects with different textures. 
Julesz introduced textons \shortcite{Julesz:81} as a way to represent texture via second order statistics of small patches.
Various filter banks can be used for texture representation \cite{randen1999filtering}, \eg Gabor filters \cite{gabor1946theory}. Also, hierarchical filter responses were used with great success for texture synthesis \cite{BonetV98,Heeger:1995}. All these filters are fixed and not learned from data.
In contrast, we learn the embedding by analyzing the distribution of all natural patches, thus avoiding the bias of hand crafted features.

The idea of representing a patch by its pixel values (without attempting dimensionality reduction) had success in various applications \cite{VarmaZ03}, see Barnes and Zhang \shortcite{Barnes2017} for a survey. In \Cref{sec:results}, we compare our method against a raw pixel descriptor.

\v{Z}bontar and LeCun \shortcite{Zbontar:2016:SMT:2946645.2946710} train a CNN to do stereo matching on image patches.
Simo-Serra~\etal~\shortcite{simo2015discriminative} learn SIFT-like descriptors
using a Siamese network. 
Both of these methods focus on invariance to viewpoint changes, whereas we aim to learn invariance to fluctuations in patch appearance of similar objects.

PatchNet~\cite{Hu:2013:PPI:2508363.2508381} introduces a compact and hierarchical representation
of image regions. It uses raw L*a*b* pixel values to
represent patches, which we compare against in \Cref{sec:results}.
PatchTable \cite{Barnes:2015:PEP:2809654.2766934} proposes an efficient approximate nearest neighbor (ANN) implementation. ANN is an orthogonal and complementary task to patch representation.

Recently, deep networks were used for image region representation and segmentation. Cimpoi~\etal~\shortcite{7299007} use the last convolution layer of a convolutional neural network (CNN) as an image region descriptor. It is not suitable for patch representation, as it will produce a 65K dimensional vector \emph{per patch.} Fully Convolutional Networks (FCNs) \cite{Long_2015_CVPR} prove potent for, \eg, image segmentation. We compare to FCNs in \Cref{sec:results}.

Our work is based on Patch2Vec~\shortcite{Patch2Vec}, which also uses deep networks to train a meaningful patch representation. However, in contrary to our method, Patch2Vec is a \emph{supervised} method that requires an annotated segmentation dataset for training.


The ideas of using spatial proximity in image space and temporal proximity for videos has been utilized in the past for self-supervised learning. Isola~\etal~\shortcite{DBLP:journals/corr/IsolaZKA15} utilize space and time co-occurrences to learn patch, frame and photo affinities. Wang and Gupta \shortcite{7410677} track objects in videos to generate data for a self-supervised learning scheme. Closer to our method, Doersch~\etal~(UVRL)~\shortcite{doersch2015unsupervised} train a network to predict the spatial relationship between pairs of patches, and use the patch representation to group similar visual concepts. Pathak~\etal~\shortcite{pathakCVPR16context} train a network to predict missing content based on its spatial surrounding. These methods learn the patch representation while training the network for a different task, and the embedding is provided implicitly. In our work, the network is directly trained for patch embedding. We compare our method against UVRL in \Cref{sec:results}.


Given a labeled set in a \emph{source} domain and an unlabeled set of samples in a \emph{target} domain, domain adaptation aims to generalize the classifier learned on the source domain to the target domain \cite{ben2010theory,chen2012marginalized}. It has become common practice to pre-train a classifier on a large labeled image database, such as ImageNet~\cite{russakovsky2015imagenet}, and transfer the parameters to a target domain \cite{oquab2014learning,sharif2014cnn}. 
See Patel~\etal~\shortcite{patel2015visual} for a survey of recent visual techniques. In our work, we refine our embeddings from the natural image source domain to a target domain that contains a common object. 
Unlike recent unsupervised domain adaptation techniques \cite{Ganin:2015:UDA:3045118.3045244,7410639}, in our case neither domain contains labeled data.

	\section{Patch Space Embedding}
\label{sec:method}

In this work, we take advantage of the fact that there is a strong coherence in the appearance of semantic segments in natural images. It is expected then that nearby patches have similar appearance. The correlation between spatial proximity and appearance similarity is learned and encoded in a patch space, where the Euclidean distance between two patches reflects their appearance similarity. 

The embedding patch space is learned by training a neural network using a triplet loss:

\begin{equation} \label{loss_function}
\begin{split}
&L(p_c, p_n, p_f) = \\
&max(0,{\Pnorm{f(p_c) - f(p_n)}{2}^2 - \Pnorm{f(p_c) - f(p_f)}{2}^2 + m}),
\end{split}
\end{equation}
where $ p_c, p_n, p_f$ are three patches of size $16 \times 16$ selected from a collection of natural images, such that $p_c$ is the current patch, $p_n$ is a nearby patch, $p_f$ is a distant patch, and $m$ is a margin value (set empirically to 0.2).

\begin{figure}[ht]
\centering
\includegraphics[width=3.0in]{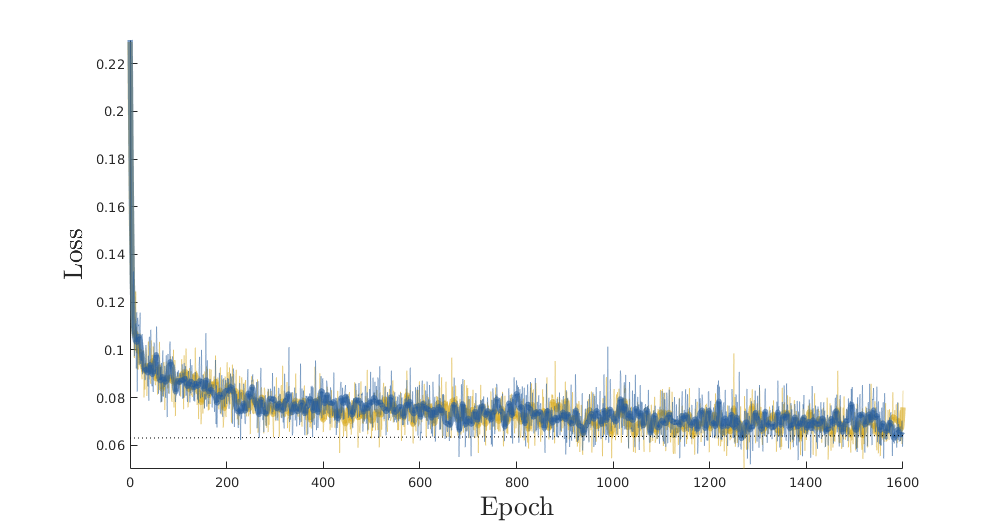}
\caption{\textbf{Network loss convergence}. The graph demonstrates the losses on the training and test data (in yellow and blue, respectively). As can be observed, the loss function is not completely stable due to the presence of outliers swatches. Nonetheless, the learning converges for both sets (starting from a loss of around $0.22$, down to $0.07$), which demonstrates the network resiliency to outliers.
}
\label{fig:losses_figure}
\end{figure}

To train our network, we utilize a large number of natural images (5000 images from the  MIT-Adobe FiveK Dataset, in our implementation) and for each image we sample six disjoint regions, referred to as swatches. Each swatch consists of a grid of nine patches. A triplet is formed by randomly picking two patches from one swatch, and one from another swatch. The assumption is that the two patches taken from the same swatch are close enough, while the third is distant. In our implementation, the distant patch is always taken from the same image. The above scheme for sampling triplets is illustrated in \Cref{fig:core_idea}, where only two swatches are illustrated, one in red and one in blue. A triplet is formed by sampling two positive patches from the red swatch, and one negative patch from the blue one. Furthermore, we adopt the principal described in \cite{Patch2Vec} that selects the "hard" examples, i.e., in each epoch, we use triplets that so far did not perform well by the network. This is expressed by the following equation:

\begin{equation}
\begin{split}
\mathcal{N}(p_c, p_n) = \{p_f \mid L(p_c, p_n, p_f) > 0 \}.
\end{split}
\end{equation}

Thus, the set $\mathcal{N}$ contains distant patches that the network embedded them within the margin $m$.
The network $f(p)$ is trained to create an embedding space that admits to the training triplets. Once trained, $f(p)$ can embed any give patch by feed-forwarding it through the network, yielding its 128D feature vector.

To cope with outliers, we incorporate a strong regularization into the network. The embedding lies only on the unit hypersphere, which prevents overfitting. The unit hypersphere provides a structure to the embedding space that is otherwise unbounded.

\begin{figure*}
	\centering
	\includegraphics[width=1.0\textwidth]{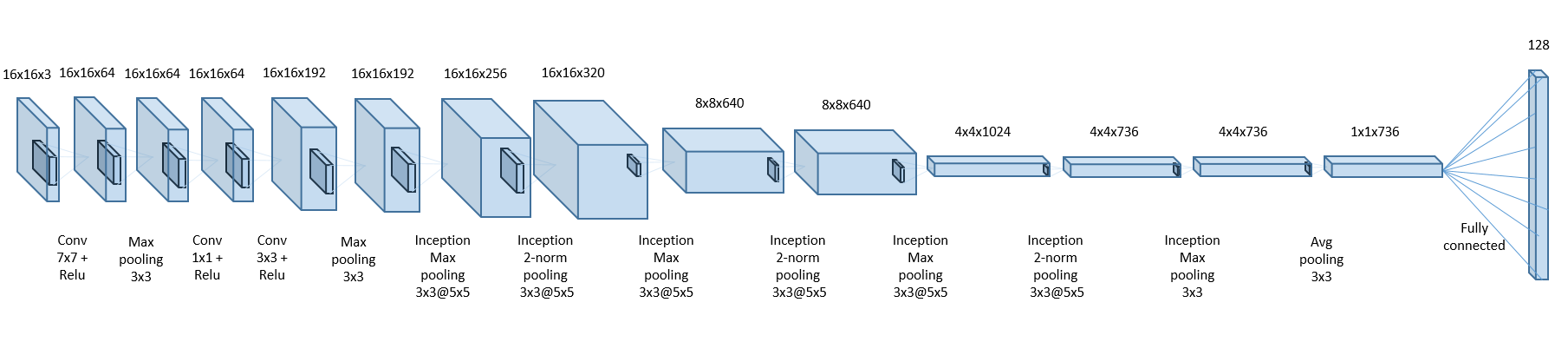}
	\caption{Our network architecture. }
	\label{fig:net_arc}
\end{figure*}

The architecture of our network is similar to the one used in \cite{Patch2Vec}, but with the required changes for supporting 16X16 patches size (see ~\Cref{fig:net_arc} for the network illustration). Note that inception layers are implemented as detailed in Szegedy~\etal~\shortcite{szegedy2015going}.

We train the network for 1600 epochs on NVIDIA GTX 1080. Training takes approximately 24 hours.
The network convergence is demonstrated in \Cref{fig:losses_figure}. As the figure illustrates, the losses on the train and test (colored in yellow and blue, respectively) are similar. This implies that our basic assumption holds and generalizes well. Furthermore, although the learning converges, the convergence is not completely stable. This may be attributed to the presence of outliers in the swatches, i.e., two patches from the same swatch but not from the same segment or two patches from different swatches but from the same segment.

\section{Domain Specialization} 
\label{sec:new_domain}

In \Cref{sec:method}, we described an unsupervised technique to embed any natural image patch onto a 128D vector. 
Given a new domain that contains a common foreground object, we can improve the embedding by fine-tuning the network, or simply training it on patches taken from the new domain. However, we can do better using the initial embedding obtained by the previously described method to generate a preliminary segmentation. We can then use these rough segments to ``supervise'' the refined embedding.

\begin{figure}[t]
\centering
\rotatebox[origin=l]{90}{  \footnotesize{Input image}}
\includegraphics[width=0.95\columnwidth]{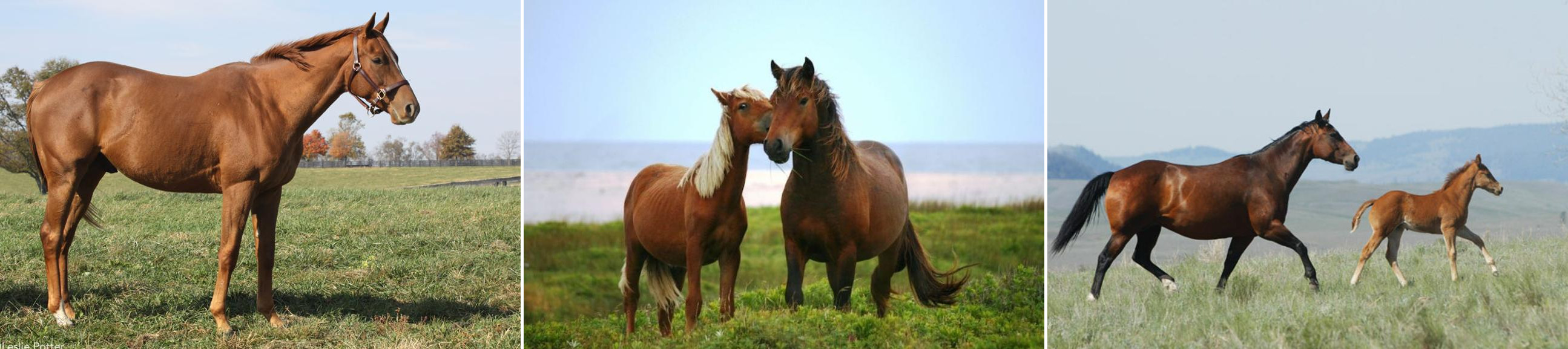} \\
\rotatebox[origin=l]{90}{\footnotesize{Output(before)}}
\includegraphics[width=0.95\columnwidth]{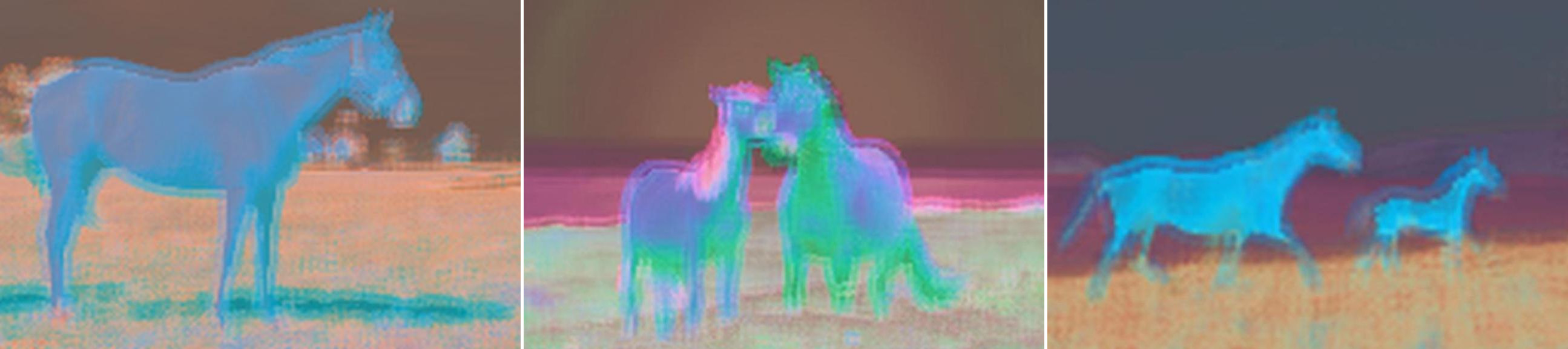} \\
\rotatebox[origin=l]{90}{\footnotesize{Guiding parts}}
\includegraphics[width=0.95\columnwidth]{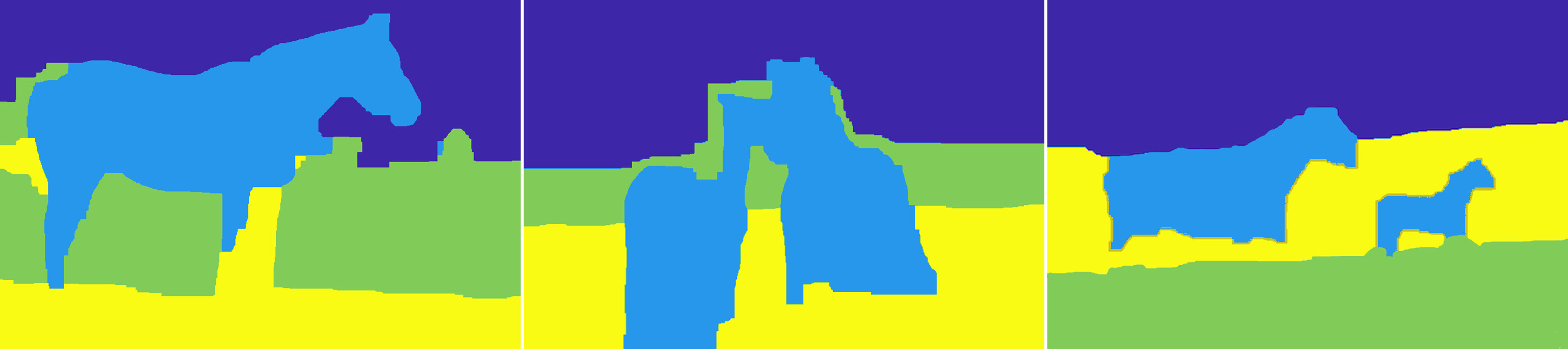} \\
\rotatebox[origin=l]{90}{\footnotesize{Output(after)}}
\includegraphics[width=0.95\columnwidth]{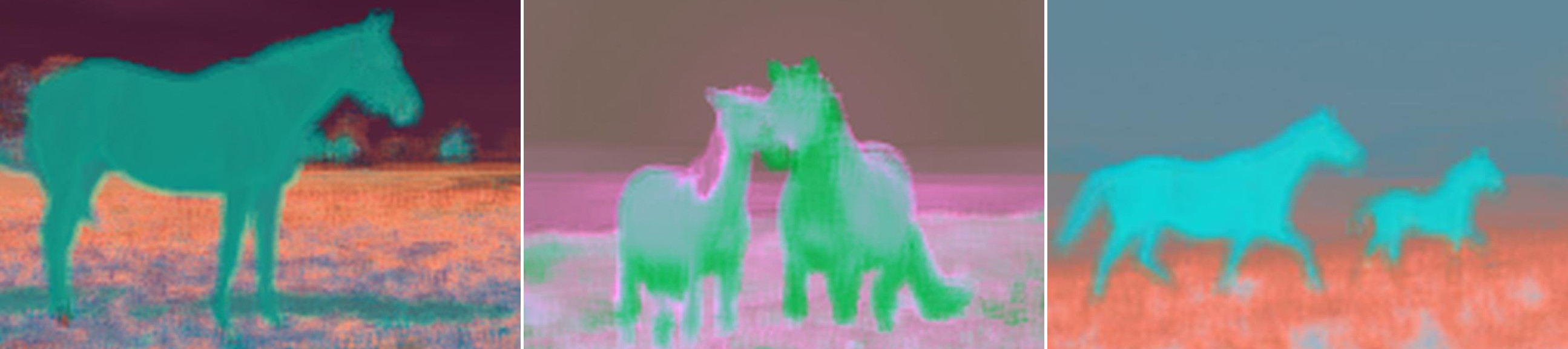} \\
\caption{Refining the embedding using self-supervision. Given a new domain that contains a common foreground object (i.e., the input images on the top), we refine our initial embedding (visualized in the second row) by automatically generating semantic guiding segments (colored in unique colors in the third row) for the training images. Our technique yields a more coherent embedding of the common object (visualized in the bottom row).  
}
\label{fig:segmentaion}
\end{figure}

To generate the rough segments, the images are first transformed using the patch embedding such that each pixel is mapped to a vector of 128D. Next, we apply k-means clustering on each of the deep images using $k = 4$ followed by a graph-cuts segmentation \cite{Bagon2006} (see the third row in \Cref{fig:segmentaion}).

These segments are then used as supervision for fine-tuning the network, where the triplets are defined based on these segments, i.e., $p_c$ and $p_n$ are taken from the same foreground segment, and $p_f$ is a patch taken from any other segment in the image. 

In our experiments, we execute the fine-tuning process only for $400$ epochs.
This process improves our embedding space and makes it much more coherent (see \Cref{tbl:auc_transfer_learning} and \Cref{fig:pca_compare}). 

 
	\section{Results}
\label{sec:results}

We performed quantitative and qualitative evaluations to analyze the performance of our embedding technique. 
The quantitative evaluation was conducted on ground truth images from the Berkeley Segmentation Dataset (BSDS500)\cite{amfm_pami2011}, which contains natural images that span a wide range of objects, as well as images from object-specific internet datasets of Rubinstein ~\etal~ \cite{Rubinstein13Unsupervised}. These object-specific datasets further enabled a quantitative evaluation of our domain specialization technique.

To quantitatively demonstrate our improved performance over previous works, we follow the measure reported by Fried~\etal~\shortcite{Patch2Vec}. We start by sampling ``same segment'' and ``different segment'' pairs of patches and calculate their distance in the embedding space. Next, for a given distance threshold, we predict that all pairs below the threshold are from the same segment, and evaluate the prediction (for all threshold values) by calculating the area under the receiver operating characteristic (ROC) curve.

\Cref{tbl:auc} contains the full comparison. Notice that \cite{Patch2Vec} is \emph{supervised}, requiring an annotated segmentation dataset. The comparison to raw RGB pixels provides a more intuitive baseline. On the other hand, the accuracy of a human annotator (bottom row in \Cref{tbl:auc}) demonstrates the problem ambiguity and a level of accuracy which can be considered ideal.
\begin{table}
  \begin{center}
    \begin{tabular}{ lcc }
      \toprule
        Method                                  & Accuracy   & Unsupervised   \\
        \midrule
        Raw pixels (RGB)                        & 0.69  & \checkmark \\
        UVRL \cite{doersch2015unsupervised}     & 0.70  & \checkmark \\
        Patch2Vec    \cite{Patch2Vec}           & 0.76  & --         \\
        Ours                                    & 0.78  & \checkmark \\[0.6ex]
        Human                                   & 0.86  & --         \\
      \bottomrule
    \end{tabular}
  \end{center}
 \vspace{1pt}
  \caption{\textbf{Patch Embedding Evaluation}. We compare our method to alternative patch representations. We report the AUC scores using l2 distance between patch
representation as means to predict if a pair of patches comes from the same segment or
not.
  }

  \label{tbl:auc}
\end{table}

\begin{figure}
	\centering
   
	\rotatebox[origin=l]{90}{  \whitetxt{ssssss}Input}
	\adjincludegraphics[width=0.23\columnwidth]{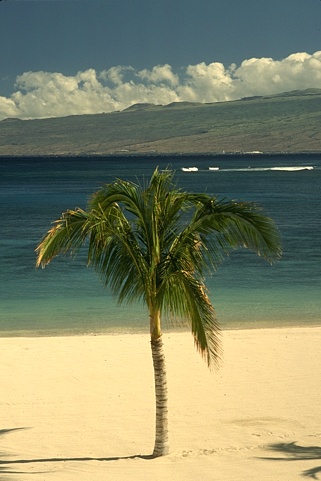}  
	\adjincludegraphics[width=0.23\columnwidth]{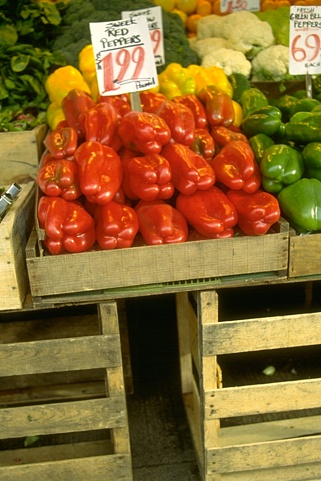}  
    \adjincludegraphics[width=0.23\columnwidth]{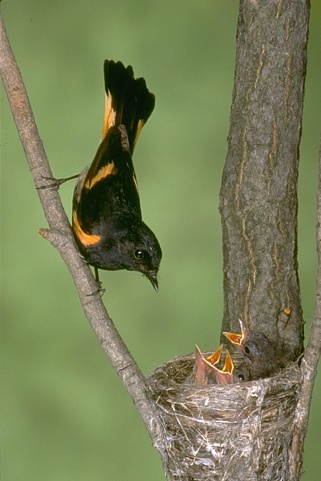} 
    \adjincludegraphics[width=0.23\columnwidth]{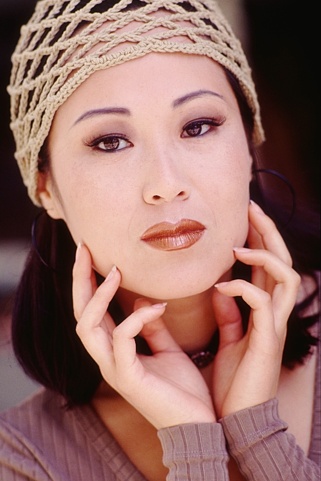} 
    \\
    
	\rotatebox[origin=l]{90}{\whitetxt{ss}Patch2Vec\protect\cite{Patch2Vec}}
	\adjincludegraphics[width=0.23\columnwidth]{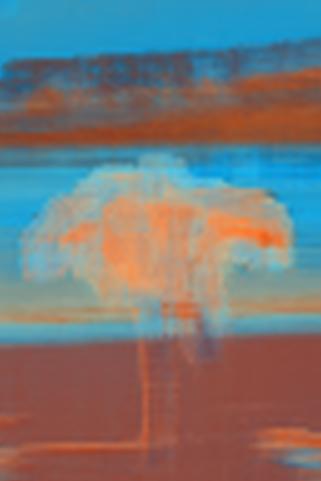}  
	\adjincludegraphics[width=0.23\columnwidth]{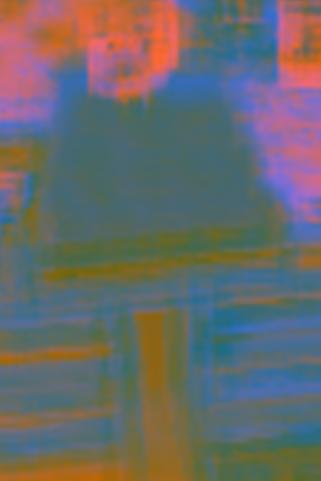}  
    \adjincludegraphics[width=0.23\columnwidth]{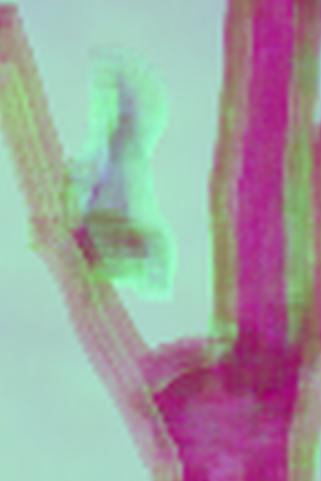} 
    \adjincludegraphics[width=0.23\columnwidth]{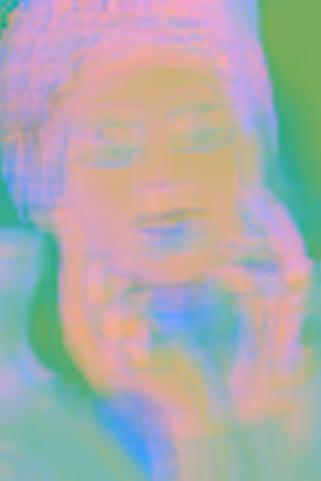} 
    \\
	\rotatebox[origin=l]{90}{\whitetxt{sssssI}Ours}
    \hspace{0.25pt}
	\adjincludegraphics[width=0.23\columnwidth]{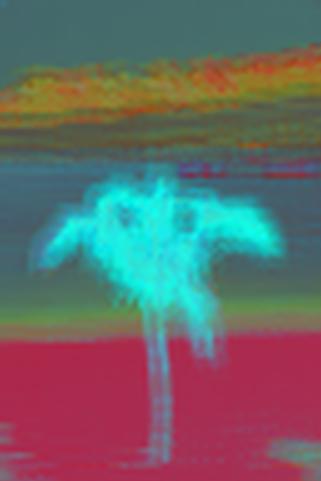}  
	\adjincludegraphics[width=0.23\columnwidth]{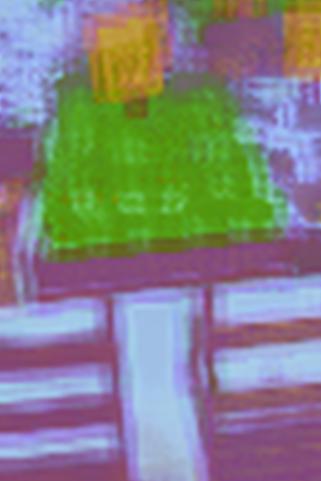}  
    \adjincludegraphics[width=0.23\columnwidth]{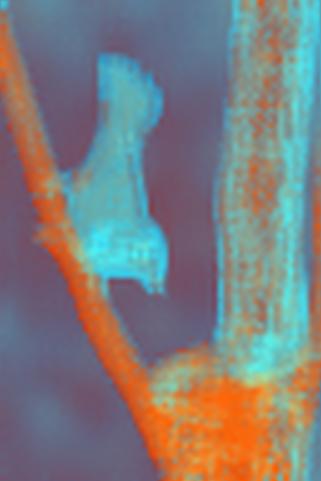} 
    \adjincludegraphics[width=0.23\columnwidth]{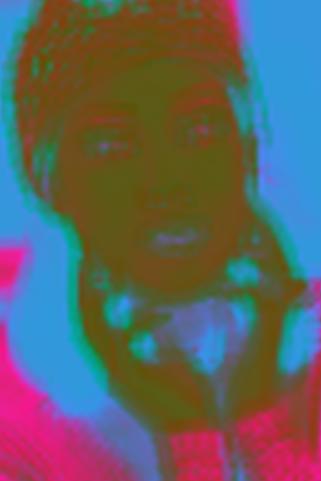} 
    \\

	\caption{\textbf{Supervised vs. unsupervised embedding technique}. The second and third rows show results of Patch2Vec \protect\cite{Patch2Vec} and our unsupervised technique, respectively. The input images, which belong to the training data of \protect\cite{Patch2Vec}, are provided on the first row. Note that although our method did not train on these images, the textures are significantly less apparent in our embeddings. This suggests that segments with similar texture are embedded to the closer locations in the embedding space.
}
	\label{fig:pca_comparison_patch2vec}
\end{figure}

To qualitatively visualize the quality of our embeddings, as previously detailed, we project the 128D vectors onto their three principle directions, which enables producing pseudo-RGB images where similar colors correspond to similar embedded points. In \Cref{fig:pca_comparison_patch2vec}, we visualize our embeddings and compare to the supervised technique of Fried~\etal~\cite{Patch2Vec} on \emph{their} training data. As the figure illustrates, our results are more coherent than the ones obtained with supervision, even though our method does not train on these images. In the supplementary material, we provide
a comparison for the \emph{full} BSDS500 dataset. Please refer to these results
for assessing the high quality of our results.

\begin{figure*}
\hspace{-5pt}
	\centering
	\rotatebox[origin=l]{90}{  \whitetxt{sss}Input}
	\adjincludegraphics[width=0.191\textwidth]{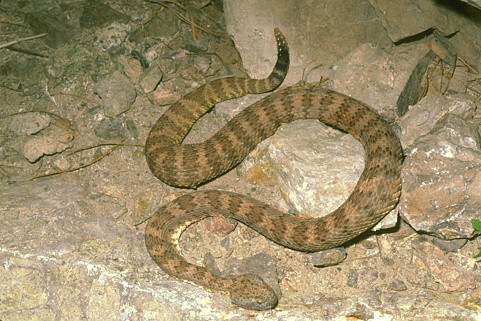}  
	\adjincludegraphics[width=0.191\textwidth]{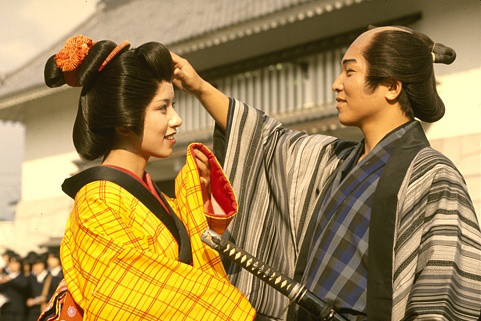}
    \adjincludegraphics[width=0.191\textwidth]{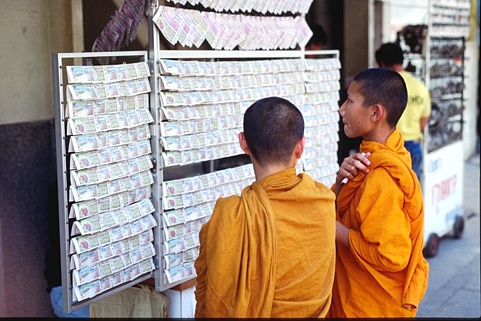}
    \adjincludegraphics[width=0.191\textwidth]{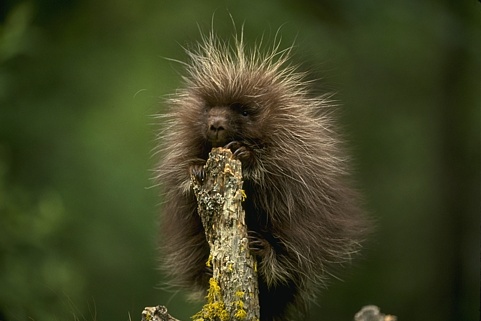}
    \adjincludegraphics[width=0.191\textwidth]{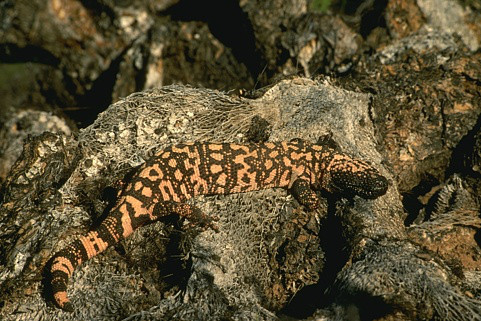} 
    \footnotesize{\whitetxt{sssssssssssssssss}} \vspace{-8pt}
    \\
        \rotatebox[origin=l]{90}{\whitetxt{ssss}UVRL\protect\cite{doersch2015unsupervised}}
	\adjincludegraphics[width=0.192\textwidth]{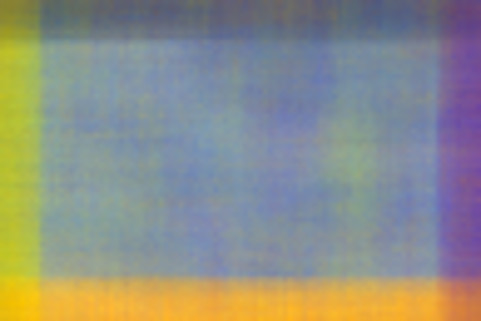}  
	\adjincludegraphics[width=0.192\textwidth]{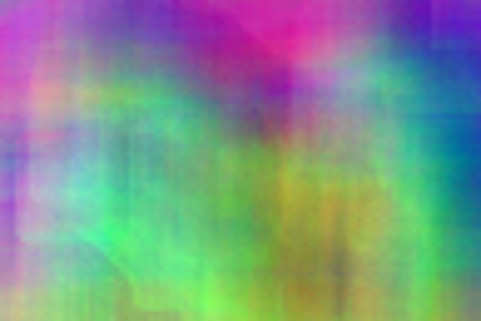}
    \adjincludegraphics[width=0.192\textwidth]{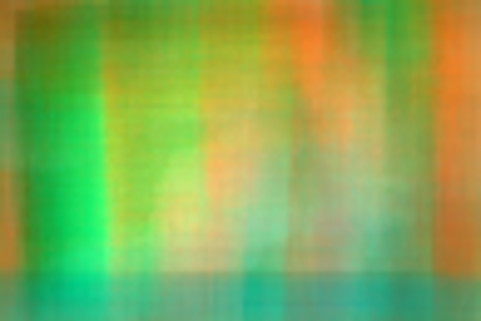}
    \adjincludegraphics[width=0.192\textwidth]{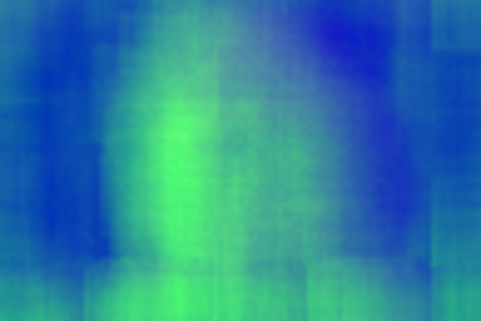}
    \adjincludegraphics[width=0.192\textwidth]{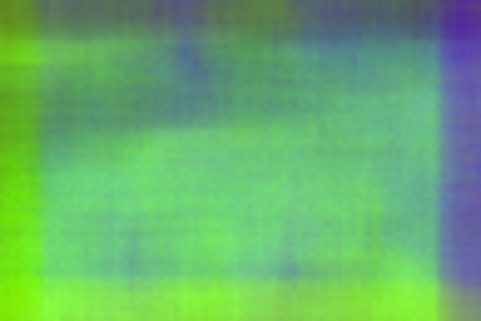} 
    \footnotesize{\whitetxt{sssssssssssssssss}} \vspace{-8pt}
    \\
	\rotatebox[origin=l]{90}{\whitetxt{ss}UVRL\protect\cite{doersch2015unsupervised}++}
	\adjincludegraphics[width=0.192\textwidth]{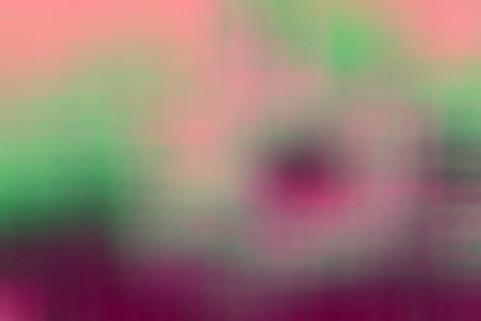}  
	\adjincludegraphics[width=0.192\textwidth]{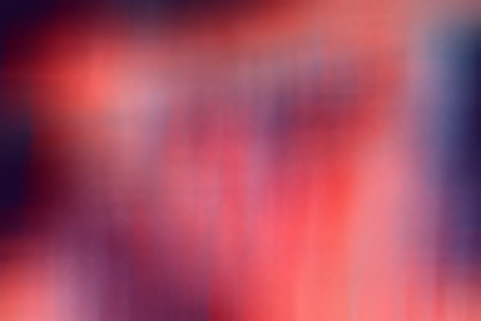}
    \adjincludegraphics[width=0.192\textwidth]{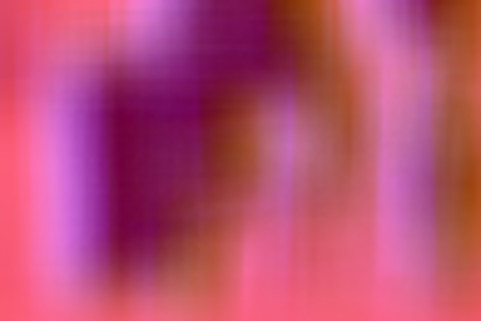}
    \adjincludegraphics[width=0.192\textwidth]{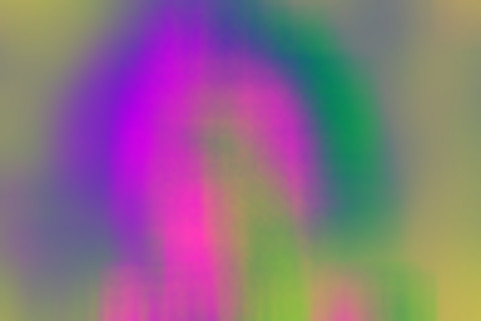}
    \adjincludegraphics[width=0.192\textwidth]{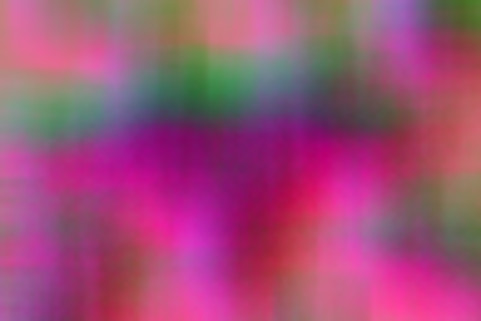} 
       \footnotesize{\whitetxt{sssssssssssssssss}} \vspace{-8pt}
       \\
     \rotatebox[origin=l]{90}{\whitetxt{Issss}Ours}
	\adjincludegraphics[width=0.192\textwidth]{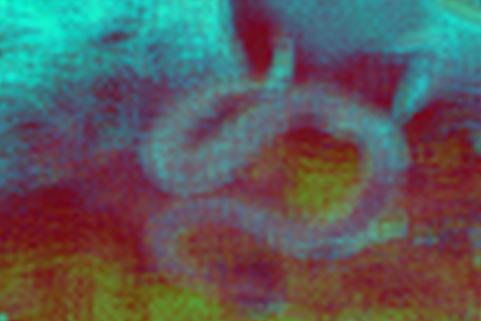}  
	\adjincludegraphics[width=0.192\textwidth]{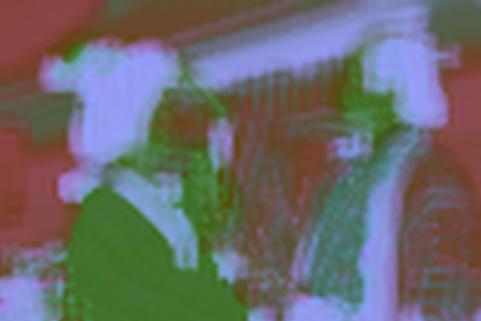}
    \adjincludegraphics[width=0.192\textwidth]{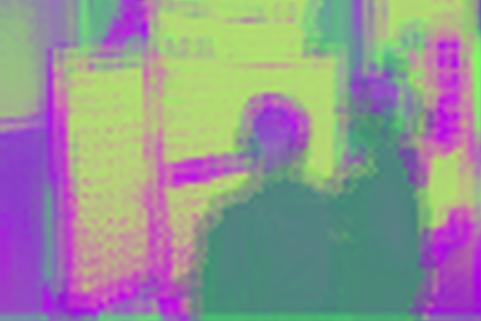}
    \adjincludegraphics[width=0.192\textwidth]{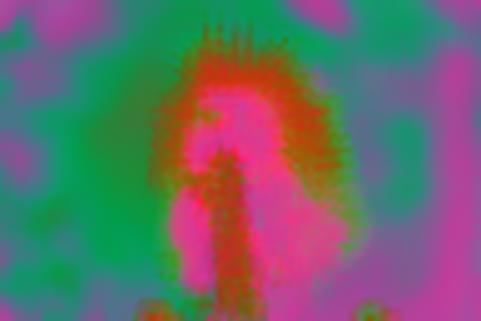}
    \adjincludegraphics[width=0.192\textwidth]{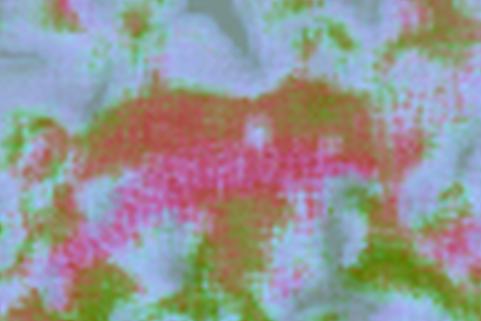} \\
	\caption{\textbf{Comparison between our embedding to one inferred by deep representations}. We compare to UVRL\protect\cite{doersch2015unsupervised} using their pre-trained weights (second row) and also by retraining them on BSDS500 (third row). As demonstrated above, our technique maps pixels from similar regions to closer values.  
    }
	\label{fig:pca_comparison_unsupervised}
\end{figure*}

In \Cref{fig:pca_comparison_unsupervised}, we compare to Doersch~\etal~\cite{doersch2015unsupervised}, where the patch representation can also be obtained without supervision. For comparison purposes, we use both their pre-trained weights and the weights
retrained on BSDS500. We use their fc6 layer which performed the
best in our tests. Unlike our embeddings, their method does not produce
similar embeddings, which are visualized by similar colors in the figure, for pixels of the same region.

\begin{table}
  \begin{center}
    \begin{tabular}{ lcc }
      \toprule
        Method                                  & Accuracy  \\
        \midrule
        Horses\\
        \midrule
        Baseline (before fine-tuning)           & 0.68  \\
        Fine-tuned (training)   		        & 0.71  \\
        Fine-tuned (testing)    		        & 0.70  \\
        Fine-tuned+self-supervision (training) 	& 0.72  \\
        Fine-tuned+self-supervision (testing)   & 0.72  \\
        \midrule
        Airplanes\\
        \midrule
        Baseline (before fine-tuning)           & 0.623  \\
        Fine-tuned (training)   		        & 0.640  \\
        Fine-tuned (testing)    		        & 0.656  \\
        Fine-tuned+self-supervision (training) 	& 0.662  \\
        Fine-tuned+self-supervision (testing)   & 0.672  \\
        \midrule
        Cars\\
        \midrule
        Baseline (before fine-tuning)           & 0.651  \\
        Fine-tuned (training)   		        & 0.655  \\
        Fine-tuned (testing)    		        & 0.653  \\
        Fine-tuned+self-supervision (training) 	& 0.670  \\
        Fine-tuned+self-supervision (testing)   & 0.670  \\
      \bottomrule
    \end{tabular}
  \end{center}
 \vspace{1pt}
  \caption{\textbf{Domain Specialization Evaluation}. Above, we report the AUC scores on the object-specific datasets provided by \cite{Rubinstein13Unsupervised} before fine-tuning the network (baseline), after fine-tuning the network on patches from the dataset (fine-tune) and after fine-tuning the network using our self-supervision technique (fine-tuned+self-supervision).
  }
  \label{tbl:auc_transfer_learning}
\end{table}

To evaluate our domain specialization technique, we fine-tune our network in two ways: First, we retrain the weights by simply training it on patches taken from the object-specific datasets of Rubinstein~\etal~ \cite{Rubinstein13Unsupervised}. The second option is the one we describe in \Cref{sec:new_domain}, where we use self-supervision to refine the results in the new domain. 

In \Cref{tbl:auc_transfer_learning}, we report the AUC scores in both settings. Since the ground-truth for these datasets contains only foreground-background segmentation (and not a segment for each semantic object), the AUC measure required a slight adjustment. As the background may contain many unrelated segments, we sample ``same segment'' pairs only from the foreground. 
As validated in \Cref{tbl:auc_transfer_learning}, our method successfully learns and adjusts to the new domain. Moreover, our self-supervision scheme further boosts the performance. 

\begin{figure*}[ht]
	\centering
    
    \rotatebox[origin=l]{90}{\whitetxt{}\small{Input}}
    \hspace{-4pt}
	\includegraphics[width=0.97\textwidth]{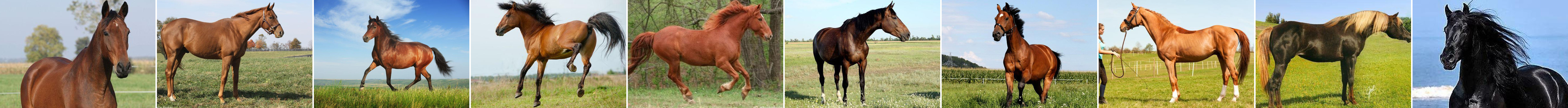} \\
    \rotatebox[origin=l]{90}{\whitetxt{}\small{Initial}}
    \includegraphics[width=0.97\textwidth]{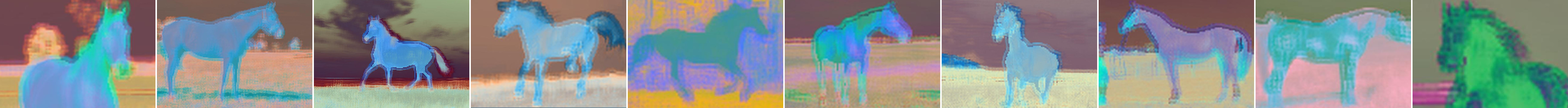} \\
     \rotatebox[origin=l]{90}{\small{Refined}}
      \includegraphics[width=0.97\textwidth]{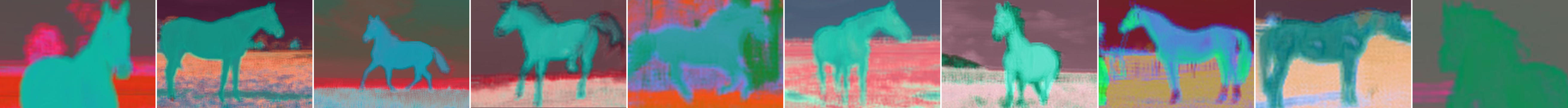}
	\caption{\textbf{Refining the embeddings to the \textsc{Horse} domain}. Above, we demonstrate the embeddings before and after the domain specialization stage. As illustrated in the figure and quantitatively stated in \Cref{tbl:auc_transfer_learning},  the embeddings of the objects (e.g., the horses) are more coherent after the refinement.  }
	\label{fig:pca_compare}
\end{figure*}

In \Cref{fig:pca_compare}, we qualitatively demonstrate the improvement over samples belonging to the \textsc{Horse}, \textsc{Car} and \textsc{Airplane} datasets, half of them belong to the training set and the other half belong to the test set. Since we could not tell the samples apart, in the figure they are mixed together. As the figure illustrates, the colors, and thereby the embeddings, of the horses parts are more compatible and in general more homogeneous. For more results, see the supplementary material.

\ignorethis{
\subsection{Embedding results}
Our first step in the algorithm is the embedding, as details above we train a CNN on 2.5M created from 5000 images taken from MIT-Adobe FiveK Dataset.
We first demonstrate the embedding by performing PCA and project the embedding vectors on the three largest principal components, see \Cref{fig:pca_comparison_patch2vec}.

Also we compare our method to~\cite{Patch2Vec}  method using their benchmark. see \Cref{tbl:auc} for comparison results.

\subsection{Segmentation results}
By comparing the results to~\cite{Patch2Vec} we already know that we achieve better then working on raw pixel data, As mention in ~\cite{Patch2Vec} working on raw pixels got AUC of 0.73 on ground-truth data, from Berkeley Segmentation Dataset (BSDS500). We also compare our embedding space vs RGB raw data as the embedding space go through all our algorithm pipeline on new and unlabeled domain, see \Cref{fig:rgb_vs_ours_results}.

\subsection{Foreground results}
We also demonstrate our algorithm  in order to perform foreground extraction. So we take about 1000 horses images from Google, and perform our fine-tuning on it, as described in \Cref{new_domain}, see \Cref{fig:full_pipeline_results}. 
}

\ignorethis{
\begin{figure}
	\centering
	\includegraphics[width=1.5in]{images/rgb_vs_ours/ours_after_domain_transfer/Google_horse_0000_4_}
	\includegraphics[width=1.5in]{images/rgb_vs_ours/rgb/Google_horse_0000_5_}
	\includegraphics[width=1.5in]{images/rgb_vs_ours/ours_after_domain_transfer/Google_horse_0001_4_}
	\includegraphics[width=1.5in]{images/rgb_vs_ours/rgb/Google_horse_0001_5_}
	\includegraphics[width=1.5in]{images/rgb_vs_ours/ours_after_domain_transfer/Google_horse_0004_4_}
	\includegraphics[width=1.5in]{images/rgb_vs_ours/rgb/Google_horse_0004_5_}
	\includegraphics[width=1.5in]{images/rgb_vs_ours/ours_after_domain_transfer/Google_horse_0005_4_}
	\includegraphics[width=1.5in]{images/rgb_vs_ours/rgb/Google_horse_0005_6_}
	\includegraphics[width=1.5in]{images/rgb_vs_ours/ours_after_domain_transfer/Google_horse_0016_3_}
	\includegraphics[width=1.5in]{images/rgb_vs_ours/rgb/Google_horse_0016_4_}
	\caption{Embedding \dcor{or segmentation???} results on RBG vs. ours: On the left column we see our segmentation results, on the right column the same segmentation applied directly on RGB channels \dcor{the caption was not clear and I might miss the whole point}}
	\label{fig:rgb_vs_ours_results}
\end{figure}
}

\ignorethis{
\begin{figure}[ht]
	\centering
	\includegraphics[width=3.0in]{images/Google_horse_0000}
	\includegraphics[width=3.0in]{images/Google_horse_0005}
	\includegraphics[width=3.0in]{images/Google_horse_0007}
	\includegraphics[width=3.0in]{images/Google_horse_0041}
	\includegraphics[width=3.0in]{images/Google_horse_0048}
	\caption{Full algorithm pipeline results: From left to right - Original image, PCA on the fine tuned embedding space, GraphCut segmentation \protect\cite{Bagon2006} with k=4, Selected foreground segments combination, Add the results segment on the original image.}
	\label{fig:full_pipeline_results}
\end{figure}
}
	\section{Summary, Limitation, and Future Work}

We presented an unsupervised patch embedding technique, where the network learns to map natural image patches into 128D codes such that the L2 metric reflects their similarity. We showed that the triplet loss that we use to train the network explicitly for embedding outperforms other embeddings that are inferred by deep representations learned for other tasks or designed specifically to learn similarities between patches. Generally speaking, learning to embed by a network has its limitations as it is applied on the patch level. Feed-forwarding patches in network is a computationally-intensive task, and analyzing an image as a series of patches is time consuming. Parallel analysis of multitude of patch, possibly overlapping ones, can significantly accelerate the process.

To refine the performance and transfer the learning into a new domain, we utilize the embedding obtained by trained network as self-supervision. The embedded image is segmented by some naive method, to yield a rough segmentation. As demonstrated, these segments although imperfect, can successfully supervise the refinement of the network for the given new domain. However, we believe this can be further improved by using more advanced segmentation methods. In the future, we also want to consider conservative segmentation, where the segments may not necessarily cover the entire image, excluding regions with low confidence. 

Furthermore, in the future, we would like to utilize our embedding technique to advance segmentation and foreground extraction methods. In particular, we would like to analyze large sets of embedded images, aiming to co-segment the common foreground of a weakly supervised set. We believe that the common foreground object can serve as a self-supervision to further improve the embedding performance.

\ignorethis{
We want to improve our segmentation and foreground extraction process, one step was performing manually, as described on \Cref{sec:foreground_extract}, we want to automate it by using the most tight cluster and using best buddies principle, (see \cite{averbuch2015distilled}). We believe that distilling the tightest segment will results the most common foreground in the domain.
We also want to show that we can support various domains, and show our results on different domain aspect, like domains without any salient foreground. 

Our algorithm is iterative, and we perform only one step of it iteration, in will be interesting to see what happed after several iterations, furthermore it will be interesting to understand if we can add ours post processing  algorithm to the end of the network as a hyper parameter step, and see if it can improve results, in other tasks.
}


	\ifCLASSOPTIONcaptionsoff
	\newpage
	\fi

	\bibliographystyle{IEEEtran}
	\bibliography{bibliography}

\begin{thebibliography}{10}
\providecommand{\url}[1]{#1}
\csname url@samestyle\endcsname
\providecommand{\newblock}{\relax}
\providecommand{\bibinfo}[2]{#2}
\providecommand{\BIBentrySTDinterwordspacing}{\spaceskip=0pt\relax}
\providecommand{\BIBentryALTinterwordstretchfactor}{4}
\providecommand{\BIBentryALTinterwordspacing}{\spaceskip=\fontdimen2\font plus
\BIBentryALTinterwordstretchfactor\fontdimen3\font minus
  \fontdimen4\font\relax}
\providecommand{\BIBforeignlanguage}[2]{{%
\expandafter\ifx\csname l@#1\endcsname\relax
\typeout{** WARNING: IEEEtran.bst: No hyphenation pattern has been}%
\typeout{** loaded for the language `#1'. Using the pattern for}%
\typeout{** the default language instead.}%
\else
\language=\csname l@#1\endcsname
\fi
#2}}
\providecommand{\BIBdecl}{\relax}
\BIBdecl

\bibitem{matviychuk2015exploring}
Y.~Matviychuk and S.~M. Hughes, ``Exploring the manifold of image patches,''
  \emph{Proceedings of Bridges}, pp. 339--342, 2015.

\bibitem{shi2007mapping}
K.~Shi and S.-C. Zhu, ``Mapping natural image patches by explicit and implicit
  manifolds,'' in \emph{CVPR}.\hskip 1em plus 0.5em minus 0.4em\relax IEEE,
  2007.

\bibitem{Patch2Vec}
O.~Fried, S.~Avidan, and D.~Cohen-Or, ``Patch2vec: Globally consistent image
  patch representation,'' \emph{Computer Graphics Forum (Proc. Pacific
  Graphics)}, 2017.

\bibitem{doersch2015unsupervised}
C.~Doersch, A.~Gupta, and A.~A. Efros, ``Unsupervised visual representation
  learning by context prediction,'' in \emph{ICCV}, 2015.

\bibitem{Julesz:81}
B.~Julesz, ``{Textons, the elements of texture perception, and their
  interactions},'' \emph{Nature}, vol. 290, no. 5802, pp. 91--97, Mar. 1981.

\bibitem{randen1999filtering}
T.~Randen and J.~H. Husoy, ``Filtering for texture classification: A
  comparative study,'' \emph{IEEE Transactions on pattern analysis and machine
  intelligence}, vol.~21, no.~4, pp. 291--310, 1999.

\bibitem{gabor1946theory}
D.~Gabor, ``Theory of communication. part 1: The analysis of information,''
  \emph{Journal of the Institution of Electrical Engineers-Part III: Radio and
  Communication Engineering}, vol.~93, no.~26, pp. 429--441, 1946.

\bibitem{BonetV98}
J.~S.~D. Bonet and P.~A. Viola, ``Texture recognition using a non-parametric
  multi-scale statistical model,'' in \emph{CVPR}, 1998, pp. 641--647.

\bibitem{Heeger:1995}
D.~J. Heeger and J.~R. Bergen, ``Pyramid-based texture analysis/synthesis,'' in
  \emph{SIGGRAPH}, 1995, pp. 229--238.

\bibitem{VarmaZ03}
M.~Varma and A.~Zisserman, ``Texture classification: Are filter banks
  necessary?'' in \emph{CVPR}, 2003, pp. 691--698.

\bibitem{Barnes2017}
C.~Barnes and F.-L. Zhang, ``A survey of the state-of-the-art in patch-based
  synthesis,'' \emph{Computational Visual Media}, vol.~3, no.~1, pp. 3--20, Mar
  2017.

\bibitem{Zbontar:2016:SMT:2946645.2946710}
\BIBentryALTinterwordspacing
J.~\v{Z}bontar and Y.~LeCun, ``Stereo matching by training a convolutional
  neural network to compare image patches,'' \emph{J. Mach. Learn. Res.},
  vol.~17, no.~1, pp. 2287--2318, Jan. 2016. [Online]. Available:
  \url{http://dl.acm.org/citation.cfm?id=2946645.2946710}
\BIBentrySTDinterwordspacing

\bibitem{simo2015discriminative}
E.~Simo-Serra, E.~Trulls, L.~Ferraz, I.~Kokkinos, P.~Fua, and F.~Moreno-Noguer,
  ``Discriminative learning of deep convolutional feature point descriptors,''
  in \emph{Proceedings of the IEEE International Conference on Computer
  Vision}, 2015, pp. 118--126.

\bibitem{Hu:2013:PPI:2508363.2508381}
\BIBentryALTinterwordspacing
S.-M. Hu, F.-L. Zhang, M.~Wang, R.~R. Martin, and J.~Wang, ``Patchnet: A
  patch-based image representation for interactive library-driven image
  editing,'' \emph{ACM Trans. Graph.}, 2013. [Online]. Available:
  \url{http://doi.acm.org/10.1145/2508363.2508381}
\BIBentrySTDinterwordspacing

\bibitem{Barnes:2015:PEP:2809654.2766934}
\BIBentryALTinterwordspacing
C.~Barnes, F.-L. Zhang, L.~Lou, X.~Wu, and S.-M. Hu, ``Patchtable: Efficient
  patch queries for large datasets and applications,'' \emph{ACM Trans.
  Graph.}, vol.~34, no.~4, pp. 97:1--97:10, Jul. 2015. [Online]. Available:
  \url{http://doi.acm.org/10.1145/2766934}
\BIBentrySTDinterwordspacing

\bibitem{7299007}
M.~Cimpoi, S.~Maji, and A.~Vedaldi, ``Deep filter banks for texture recognition
  and segmentation,'' in \emph{CVPR}, June 2015, pp. 3828--3836.

\bibitem{Long_2015_CVPR}
J.~Long, E.~Shelhamer, and T.~Darrell, ``Fully convolutional networks for
  semantic segmentation,'' in \emph{CVPR}, June 2015.

\bibitem{DBLP:journals/corr/IsolaZKA15}
\BIBentryALTinterwordspacing
P.~Isola, D.~Zoran, D.~Krishnan, and E.~H. Adelson, ``Learning visual groups
  from co-occurrences in space and time,'' \emph{CoRR}, vol. abs/1511.06811,
  2015. [Online]. Available: \url{http://arxiv.org/abs/1511.06811}
\BIBentrySTDinterwordspacing

\bibitem{7410677}
X.~Wang and A.~Gupta, ``Unsupervised learning of visual representations using
  videos,'' in \emph{ICCV}, Dec 2015.

\bibitem{pathakCVPR16context}
D.~Pathak, P.~Kr\"ahenb\"uhl, J.~Donahue, T.~Darrell, and A.~Efros, ``Context
  encoders: Feature learning by inpainting,'' 2016.

\bibitem{ben2010theory}
S.~Ben-David, J.~Blitzer, K.~Crammer, A.~Kulesza, F.~Pereira, and J.~W.
  Vaughan, ``A theory of learning from different domains,'' \emph{Machine
  learning}, vol.~79, no.~1, pp. 151--175, 2010.

\bibitem{chen2012marginalized}
\BIBentryALTinterwordspacing
M.~Chen, Z.~Xu, K.~Q. Weinberger, and F.~Sha, ``Marginalized denoising
  autoencoders for domain adaptation,'' in \emph{Proceedings of the 29th
  International Conference on Machine Learning}, ser. ICML'12, USA, 2012, pp.
  1627--1634. [Online]. Available:
  \url{http://dl.acm.org/citation.cfm?id=3042573.3042781}
\BIBentrySTDinterwordspacing

\bibitem{russakovsky2015imagenet}
O.~Russakovsky, J.~Deng, H.~Su, J.~Krause, S.~Satheesh, S.~Ma, Z.~Huang,
  A.~Karpathy, A.~Khosla, M.~Bernstein \emph{et~al.}, ``Imagenet large scale
  visual recognition challenge,'' \emph{International Journal of Computer
  Vision}, vol. 115, no.~3, pp. 211--252, 2015.

\bibitem{oquab2014learning}
M.~Oquab, L.~Bottou, I.~Laptev, and J.~Sivic, ``Learning and transferring
  mid-level image representations using convolutional neural networks,'' in
  \emph{Proceedings of the IEEE conference on computer vision and pattern
  recognition}, 2014, pp. 1717--1724.

\bibitem{sharif2014cnn}
A.~Sharif~Razavian, H.~Azizpour, J.~Sullivan, and S.~Carlsson, ``Cnn features
  off-the-shelf: an astounding baseline for recognition,'' in \emph{Proceedings
  of the IEEE conference on computer vision and pattern recognition workshops},
  2014, pp. 806--813.

\bibitem{patel2015visual}
V.~M. Patel, R.~Gopalan, R.~Li, and R.~Chellappa, ``Visual domain adaptation: A
  survey of recent advances,'' \emph{IEEE signal processing magazine}, vol.~32,
  no.~3, pp. 53--69, 2015.

\bibitem{Ganin:2015:UDA:3045118.3045244}
\BIBentryALTinterwordspacing
Y.~Ganin and V.~Lempitsky, ``Unsupervised domain adaptation by
  backpropagation,'' in \emph{Proceedings of the 32Nd International Conference
  on International Conference on Machine Learning - Volume 37}, ser.
  ICML'15.\hskip 1em plus 0.5em minus 0.4em\relax JMLR.org, 2015, pp.
  1180--1189. [Online]. Available:
  \url{http://dl.acm.org/citation.cfm?id=3045118.3045244}
\BIBentrySTDinterwordspacing

\bibitem{7410639}
E.~Kodirov, T.~Xiang, Z.~Fu, and S.~Gong, ``Unsupervised domain adaptation for
  zero-shot learning,'' in \emph{ICCV}, Dec 2015.

\bibitem{szegedy2015going}
C.~Szegedy, W.~Liu, Y.~Jia, P.~Sermanet, S.~Reed, D.~Anguelov, D.~Erhan,
  V.~Vanhoucke, and A.~Rabinovich, ``Going deeper with convolutions,'' in
  \emph{Proceedings of the IEEE conference on computer vision and pattern
  recognition}, 2015, pp. 1--9.

\bibitem{Bagon2006}
\BIBentryALTinterwordspacing
S.~Bagon, ``Matlab wrapper for graph cut,'' December 2006. [Online]. Available:
  \url{http://www.wisdom.weizmann.ac.il/~bagon}
\BIBentrySTDinterwordspacing

\bibitem{amfm_pami2011}
\BIBentryALTinterwordspacing
P.~Arbelaez, M.~Maire, C.~Fowlkes, and J.~Malik, ``Contour detection and
  hierarchical image segmentation,'' \emph{IEEE Trans. Pattern Anal. Mach.
  Intell.}, vol.~33, no.~5, pp. 898--916, May 2011. [Online]. Available:
  \url{http://dx.doi.org/10.1109/TPAMI.2010.161}
\BIBentrySTDinterwordspacing

\bibitem{Rubinstein13Unsupervised}
M.~Rubinstein, A.~Joulin, J.~Kopf, and C.~Liu, ``Unsupervised joint object
  discovery and segmentation in internet images,'' \emph{IEEE Conf. on Computer
  Vision and Pattern Recognition (CVPR)}, June 2013.

\end{thebibliography}


\begin{thebibliography}{10}
\providecommand{\url}[1]{#1}
\csname url@samestyle\endcsname
\providecommand{\newblock}{\relax}
\providecommand{\bibinfo}[2]{#2}
\providecommand{\BIBentrySTDinterwordspacing}{\spaceskip=0pt\relax}
\providecommand{\BIBentryALTinterwordstretchfactor}{4}
\providecommand{\BIBentryALTinterwordspacing}{\spaceskip=\fontdimen2\font plus
\BIBentryALTinterwordstretchfactor\fontdimen3\font minus
  \fontdimen4\font\relax}
\providecommand{\BIBforeignlanguage}[2]{{%
\expandafter\ifx\csname l@#1\endcsname\relax
\typeout{** WARNING: IEEEtran.bst: No hyphenation pattern has been}%
\typeout{** loaded for the language `#1'. Using the pattern for}%
\typeout{** the default language instead.}%
\else
\language=\csname l@#1\endcsname
\fi
#2}}
\providecommand{\BIBdecl}{\relax}
\BIBdecl

\bibitem{fu2007flame}
L.~Fu and E.~Medico, ``Flame, a novel fuzzy clustering method for the analysis
  of dna microarray data,'' \emph{BMC bioinformatics}, vol.~8, no.~1, p.~3,
  2007.

\bibitem{macqueen1967some}
J.~MacQueen \emph{et~al.}, ``Some methods for classification and analysis of
  multivariate observations,'' in \emph{Proceedings of the fifth Berkeley
  symposium on mathematical statistics and probability}, vol.~1, no.~14.\hskip
  1em plus 0.5em minus 0.4em\relax Oakland, CA, USA., 1967, pp. 281--297.

\bibitem{cheng1995mean}
Y.~Cheng, ``Mean shift, mode seeking, and clustering,'' \emph{Pattern Analysis
  and Machine Intelligence, IEEE Transactions on}, vol.~17, no.~8, pp.
  790--799, 1995.

\bibitem{ester1996density}
M.~Ester, H.-P. Kriegel, J.~Sander, and X.~Xu, ``A density-based algorithm for
  discovering clusters in large spatial databases with noise.'' in \emph{Kdd},
  vol.~96, no.~34, 1996, pp. 226--231.

\bibitem{jain2010data}
A.~K. Jain, ``Data clustering: 50 years beyond k-means,'' \emph{Pattern
  recognition letters}, vol.~31, no.~8, pp. 651--666, 2010.

\bibitem{karypis1999chameleon}
G.~Karypis, E.-H. Han, and V.~Kumar, ``Chameleon: Hierarchical clustering using
  dynamic modeling,'' \emph{Computer}, vol.~32, no.~8, pp. 68--75, 1999.

\bibitem{karami2014choosing}
A.~Karami and R.~Johansson, ``Choosing dbscan parameters automatically using
  differential evolution,'' \emph{International Journal of Computer
  Applications}, vol.~91, no.~7, 2014.

\bibitem{sawant2014adaptive}
K.~Sawant, ``Adaptive methods for determining dbscan parameters,''
  \emph{International Journal of Innovative Science, Engineering \&
  Technology}, vol.~1, no.~4, 2014.

\bibitem{ankerst1999optics}
M.~Ankerst, M.~M. Breunig, H.-P. Kriegel, and J.~Sander, ``Optics: ordering
  points to identify the clustering structure,'' in \emph{ACM Sigmod Record},
  vol.~28, no.~2.\hskip 1em plus 0.5em minus 0.4em\relax ACM, 1999, pp. 49--60.

\bibitem{campello2013density}
R.~J. Campello, D.~Moulavi, and J.~Sander, ``Density-based clustering based on
  hierarchical density estimates,'' in \emph{Advances in Knowledge Discovery
  and Data Mining}.\hskip 1em plus 0.5em minus 0.4em\relax Springer, 2013, pp.
  160--172.

\bibitem{hinneburg1998efficient}
A.~Hinneburg and D.~A. Keim, ``An efficient approach to clustering in large
  multimedia databases with noise,'' in \emph{KDD}, vol.~98, 1998, pp. 58--65.

\bibitem{ertoz2003finding}
L.~Ert{\"o}z, M.~Steinbach, and V.~Kumar, ``Finding clusters of different
  sizes, shapes, and densities in noisy, high dimensional data.'' in
  \emph{SDM}.\hskip 1em plus 0.5em minus 0.4em\relax SIAM, 2003, pp. 47--58.

\bibitem{maier2007cluster}
M.~Maier, M.~Hein, and U.~Von~Luxburg, ``Cluster identification in
  nearest-neighbor graphs,'' in \emph{International Conference on Algorithmic
  Learning Theory}.\hskip 1em plus 0.5em minus 0.4em\relax Springer, 2007, pp.
  196--210.

\bibitem{shimshoni2006adaptive}
I.~Shimshoni, B.~Georgescu, and P.~Meer, ``Adaptive mean shift based clustering
  in high dimensions,'' \emph{Nearest-neighbor methods in learning and vision:
  theory and practice}, pp. 203--220, 2006.

\bibitem{carreira2015review}
M.~A. Carreira-Perpin{\'a}n, ``A review of mean-shift algorithms for
  clustering,'' \emph{arXiv preprint arXiv:1503.00687}, 2015.

\bibitem{frey2007clustering}
B.~J. Frey and D.~Dueck, ``Clustering by passing messages between data
  points,'' \emph{science}, vol. 315, no. 5814, pp. 972--976, 2007.

\bibitem{afpfaq}
------, ``Affinity propagation faq,''
  http://www.psi.toronto.edu/affinitypropagation/faq.html, 2007, [Online;
  accessed 16-November-2016].

\bibitem{xia2006border}
C.~Xia, W.~Hsu, M.~L. Lee, and B.~C. Ooi, ``Border: efficient computation of
  boundary points,'' \emph{Knowledge and Data Engineering, IEEE Transactions
  on}, vol.~18, no.~3, pp. 289--303, 2006.

\bibitem{korn2000influence}
F.~Korn and S.~Muthukrishnan, ``Influence sets based on reverse nearest
  neighbor queries,'' in \emph{ACM SIGMOD Record}, vol.~29, no.~2.\hskip 1em
  plus 0.5em minus 0.4em\relax ACM, 2000, pp. 201--212.

\bibitem{zelnik2004self}
L.~Zelnik-Manor and P.~Perona, ``Self-tuning spectral clustering,'' in
  \emph{Advances in neural information processing systems}, 2004, pp.
  1601--1608.

\bibitem{van2011numpy}
S.~Van Der~Walt, S.~C. Colbert, and G.~Varoquaux, ``The numpy array: a
  structure for efficient numerical computation,'' \emph{Computing in Science
  \& Engineering}, vol.~13, no.~2, pp. 22--30, 2011.

\bibitem{scikit-learn}
F.~Pedregosa, G.~Varoquaux, A.~Gramfort, V.~Michel, B.~Thirion, O.~Grisel,
  M.~Blondel, P.~Prettenhofer, R.~Weiss, V.~Dubourg, J.~Vanderplas, A.~Passos,
  D.~Cournapeau, M.~Brucher, M.~Perrot, and E.~Duchesnay, ``Scikit-learn:
  Machine learning in {P}ython,'' \emph{Journal of Machine Learning Research},
  vol.~12, pp. 2825--2830, 2011.

\bibitem{huang2017qcc}
J.~Huang, Q.~Zhu, L.~Yang, D.~Cheng, and Q.~Wu, ``Qcc: a novel clustering
  algorithm based on quasi-cluster centers,'' \emph{Machine Learning}, vol.
  106, no.~3, pp. 337--357, 2017.

\bibitem{shah2017robust}
S.~A. Shah and V.~Koltun, ``Robust continuous clustering,'' \emph{Proceedings
  of the National Academy of Sciences}, vol. 114, no.~37, pp. 9814--9819, 2017.

\bibitem{ng2002spectral}
A.~Y. Ng, M.~I. Jordan, Y.~Weiss \emph{et~al.}, ``On spectral clustering:
  Analysis and an algorithm,'' \emph{Advances in neural information processing
  systems}, vol.~2, pp. 849--856, 2002.

\bibitem{gionis2007clustering}
A.~Gionis, H.~Mannila, and P.~Tsaparas, ``Clustering aggregation,'' \emph{ACM
  Transactions on Knowledge Discovery from Data (TKDD)}, vol.~1, no.~1, p.~4,
  2007.

\bibitem{veenman2002maximum}
C.~J. Veenman, M.~J. Reinders, and E.~Backer, ``A maximum variance cluster
  algorithm,'' \emph{Pattern Analysis and Machine Intelligence, IEEE
  Transactions on}, vol.~24, no.~9, pp. 1273--1280, 2002.

\bibitem{hubert1985comparing}
L.~Hubert and P.~Arabie, ``Comparing partitions,'' \emph{Journal of
  classification}, vol.~2, no.~1, pp. 193--218, 1985.

\bibitem{vinh2010information}
N.~X. Vinh, J.~Epps, and J.~Bailey, ``Information theoretic measures for
  clusterings comparison: Variants, properties, normalization and correction
  for chance,'' \emph{The Journal of Machine Learning Research}, vol.~11, pp.
  2837--2854, 2010.

\bibitem{lecun-mnisthandwrittendigit-2010}
\BIBentryALTinterwordspacing
Y.~LeCun and C.~Cortes, ``{MNIST} handwritten digit database,'' 2010. [Online].
  Available: \url{http://yann.lecun.com/exdb/mnist/}
\BIBentrySTDinterwordspacing

\bibitem{krizhevsky2009learning}
A.~Krizhevsky and G.~Hinton, ``Learning multiple layers of features from tiny
  images,'' 2009.

\bibitem{vedaldi15matconvnet}
A.~Vedaldi and K.~Lenc, ``Matconvnet -- convolutional neural networks for
  matlab,'' in \emph{Proceeding of the {ACM} Int. Conf. on Multimedia}, 2015.

\bibitem{jolliffe2002principal}
I.~Jolliffe, \emph{Principal component analysis}.\hskip 1em plus 0.5em minus
  0.4em\relax Wiley Online Library, 2002.

\end{thebibliography}
	
	%

	
	 \vspace{-40pt} 

	\begin{IEEEbiography}[{ \vspace{-10pt} \includegraphics[width=1.05in,height=1.05in,clip, keepaspectratio]{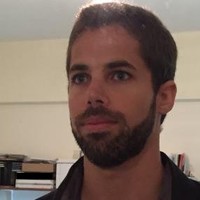}}]{Dov Danon}
		is a PhD student at the School of Computer Science, Tel-Aviv University. He received the BSc (summa cum laude) degree in computer science and mathematics from the Ben Gurion of the Negev in 2007 and the MSc degree in computer science from Tel-Aviv University in 2016. His research interests include machine learning and, in particular, unsupervised learning in image processing.
	\end{IEEEbiography}
	
	 \vspace{-40pt} 
	
	\begin{IEEEbiography}[{ \vspace{-10pt} \includegraphics[width=1.05in,height=1.05in,clip, keepaspectratio]{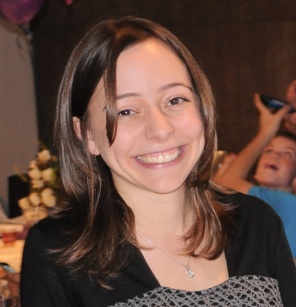}}]{Hadar Averbuch-Elor}
		is a PhD student at the School of Electrical Engineering, Tel-Aviv University and a research scientist at Amazon. She received the BSc (cum laude) degree in electrical engineering from the Technion in 2012. She worked as an computer vision algorithms developer in the defense industry from 2011 to 2015. Her research interests include computer vision and computer graphics, focusing on unstructured image collections and unsupervised techniques.
	\end{IEEEbiography}
	
	 \vspace{-40pt} 
	 
	\begin{IEEEbiography}[{ \vspace{-10pt} \includegraphics[width=1.05in,height=1.05in,clip, keepaspectratio]{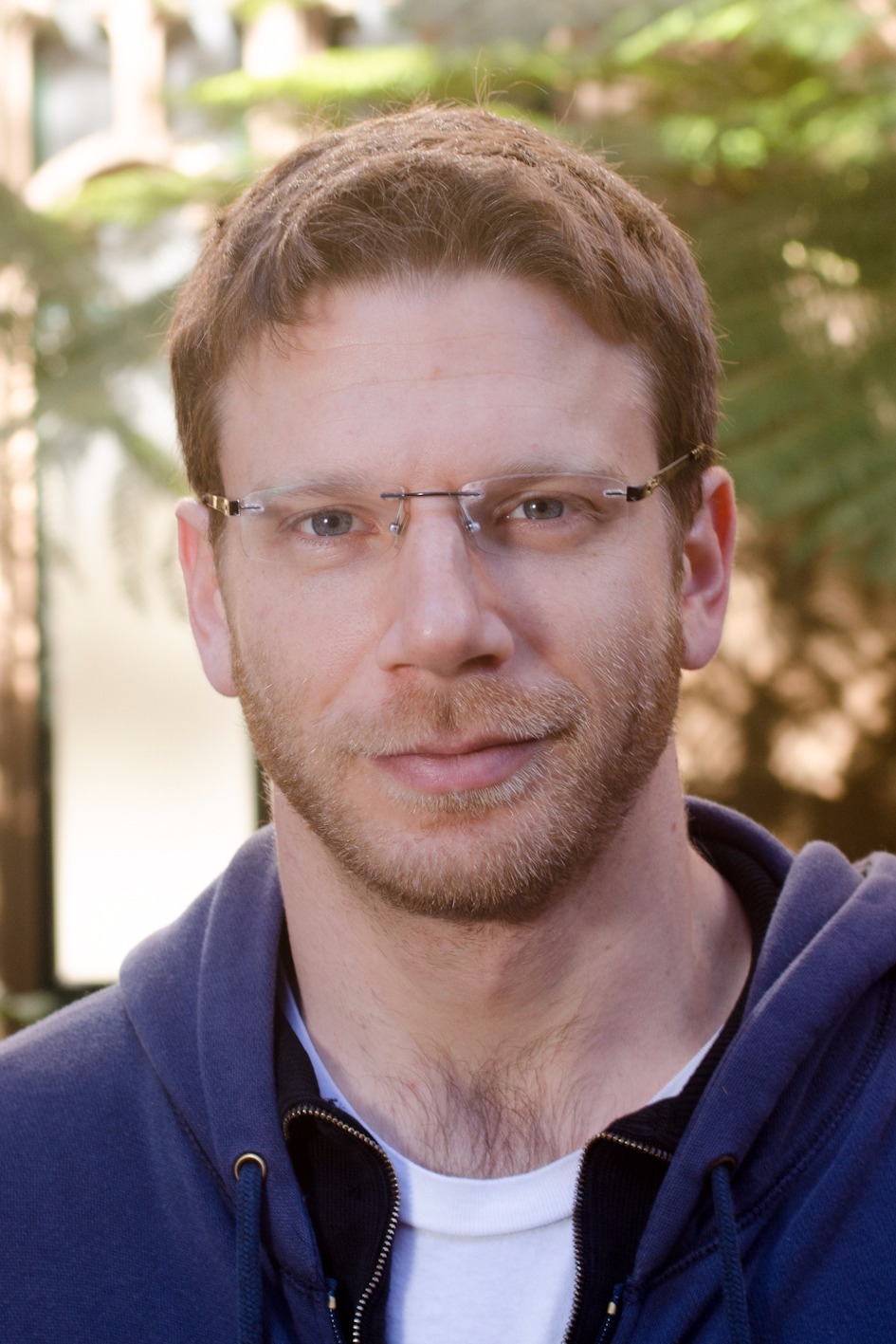}}]{Ohad Fried}
		is a postdoctoral research scholar at the school of computer science, Stanford University, and a fellow in the Brown Institute for Media Innovation. He received a B.Sc. (magna cum laude) degree in computer science and computational biology and an M.Sc. (cum laude) degree in computer science, both from the Hebrew University, in 2010 and 2012 respectively. He received a PhD from the department of computer science at Princeton University in 2017. Currently, his main interests are visual communication methods at the intersection of graphics, vision and HCI.
	\end{IEEEbiography}
	
	 \vspace{-40pt} 
	 
	\begin{IEEEbiography}[{ \vspace{-10pt} \includegraphics[width=1.05in,height=1.05in,clip, keepaspectratio]{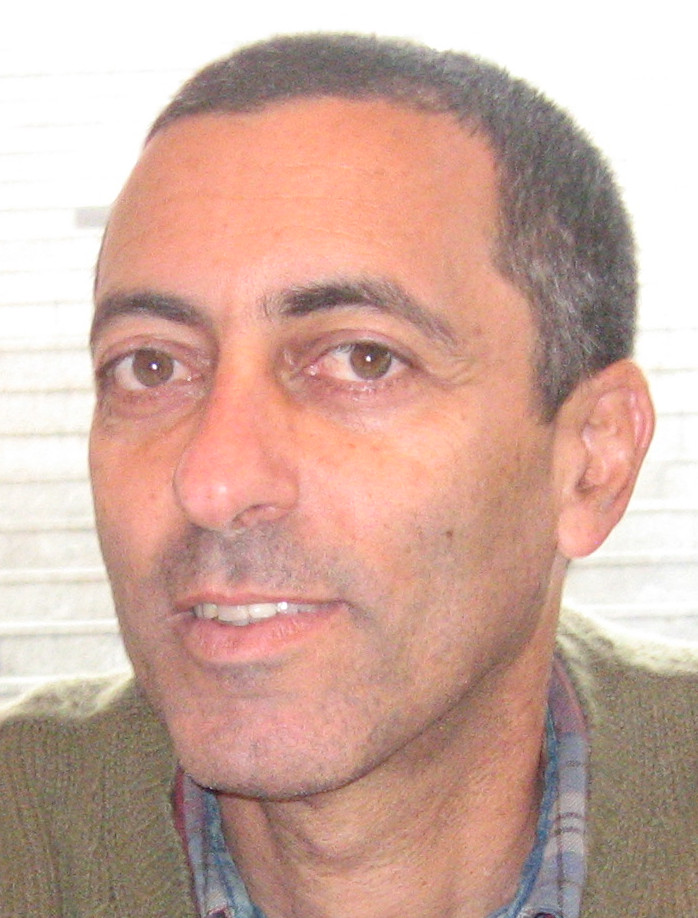}}]{Daniel Cohen-Or}
		is a professor at the School of Computer Science, Tel-Aviv University. He received the BSc (cum laude) degree in mathematics and computer Science and the MSc (cum laude) degree in computer science, both from Ben-Gurion University, in 1985 and 1986, respectively. He received the PhD from the Department of Computer Science at State University of New York at Stony Brook in 1991. He received the 2005 Eurographics Outstanding Technical Contributions Award. In 2015, he was named a Thomson Reuters Highly Cited Researcher. Currently, his main interests are in few areas: image synthesis, analysis and reconstruction, motion and transformations, shapes and surfaces.
	\end{IEEEbiography}

	
	

\end{document}